# GiAnt: A Bio-Inspired Hexapod for Adaptive Terrain Navigation and Object Detection

*Aasfee Mosharraf Bhuiyan[a], Md. Luban Mehda[a, b], Md. Thawhid Hasan Puspo, Jubayer Amin Pritom*
Department of Mechanical Engineering, Bangladesh University of Engineering & Technology, Dhaka 1000, BANGLADESH

[a]Both authors contributed equally
[b]Corresponding author: 2010137@me.buet.ac.bd

**ABSTRACT**
This paper presents the design, development and testing of GiAnt, an affordable hexapod which is inspired by the efficient motions of ants. The decision to model GiAnt after ants rather than other insects is rooted deep inside ant's natural adaptability to a variety of terrains. This bio-inspired approach gives it a significant advantage in outdoor applications having terrain flexibility along with efficient energy use. It features a lightweight 3D-printed and laser cut structure weighing 1.75 kg with dimensions of 310 mm x 200 mm x 120 mm. Its legs have been designed with a simple Single Degree of Freedom (DOF) using a link and crank mechanism. It is great for conquering challenging terrains such as grass, rocks and steep surfaces. Unlike traditional robots using 4-wheels for motion, its legged design gives super adaptability to uneven and rough surfaces. GiAnt's control system is built on Arduino, allowing manual operation. An effective way of controlling the legs of GiAnt was done by gait analysis. It can move up to 8 cm of height easily with its advanced leg positioning system. Furthermore, equipped with machine learning and image processing technology, it can identify 81 different objects in a live monitoring system. It represents a significant step towards creating accessible hexapod robots for research, exploration, surveying offering unique advantages in adaptability and control simplicity.

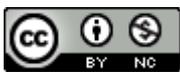

**1. Introduction**

Robotics research has advanced remarkably in recent years, with various disciplines concentrating on emulating biological locomotion to achieve improved stability, adaptability and efficiency across different terrains. Hexapod robots, inspired by multi-legged organisms like ants, have gained considerable attention for their ability to sustain mobility and stability in intricate environments. These multi-legged robots showcase enhanced stability compared to bipedal and quadrupedal designs, as each leg contributes to a wider support base, which is vital for maneuvering through challenging landscapes ([1],[2]). Emulating biological strategies, these robots strive to integrate the advantages of solid static stability and dynamic maneuverability.

Although conventional robotic mobility methods, including wheeled and bipedal systems, provide benefits in terms of speed and simplicity, their shortcomings are evident in rough, uneven conditions. Walking hexapods replicate the exceptionally stable and redundant stances observed in nature, such as those of ants, where a multi-legged posture offers resilience in rugged terrain ([3],[4]). Additionally, biologically inspired control systems enable hexapods to adapt smoothly to various terrains. Sophisticated algorithms, like terrain-adapting control and kinematic gait algorithms, enhance the capabilities of hexapods by allowing each leg to respond independently to its surroundings, thus boosting both adaptability and efficiency ([5],[6]).

Major advancements in hexapodal robotics include the Directed Angular Restitution (DAR) method, which simplifies control systems for multi-legged platforms and provides precise, continuous movement regulation. The DAR method is notable for extending trigonometric functions used in control systems, leading to smoother trajectory alignment and increased stability [3]. Additionally, recent studies emphasize the significance of feedback control systems that detect and adapt to uneven surfaces, improving navigation robustness ([5],[7]). Such innovations are crucial not only for applications in terrestrial and extraterrestrial exploration but also enhance energy efficiency, as hexapods can adjust their gait based on the terrain [8].

Research into bio-inspired robots has shown that closely mimicking the movements of ants and other insects allows robots to navigate hazardous environments with greater resilience. Ant-inspired hexapods are particularly suitable for space exploration missions, especially in Martian areas where surface irregularities entail a balanced yet flexible robotic design [2]. This design enables gait switching, facilitating seamless transitions between various movement styles, thereby offering exceptional adaptability across different environments [4].

Considering the significant progress and applications in hexapod robotics, exploring and developing ant-inspired hexapods represents an essential opportunity to create mobility solutions that excel in stability, terrain adaptability, and fault tolerance. Our initiative, aimed at creating a hexapodal robot that emulates the agility and robustness of ant movement, constitutes a meaningful contribution to the field. This innovative approach is set to tackle the complexities of challenging terrains, making it a subject of scientific and practical relevance in the field of robotics.

GiAnt stands out as a novel hexapod robot due to its innovative combination of energy efficiency, terrain adaptability and simplicity. Its design leverages a single-degree-of-freedom crank mechanism for each leg, enabling movement with just six servo motors instead of the conventional eighteen. This significantly reduces energy consumption, making it ideal for long-term missions such as exploration and surveying. It's ability to transition smoothly between walking styles and effortlessly navigate diverse terrains—including rocky surfaces and obstacles up to 8 cm high—further enhances its versatility. Additionally, the integration of simple sensors and lightweight algorithms ensures that the robot remains both affordable and accessible, making it an exceptional choice for applications requiring cost-effective, adaptive mobility.

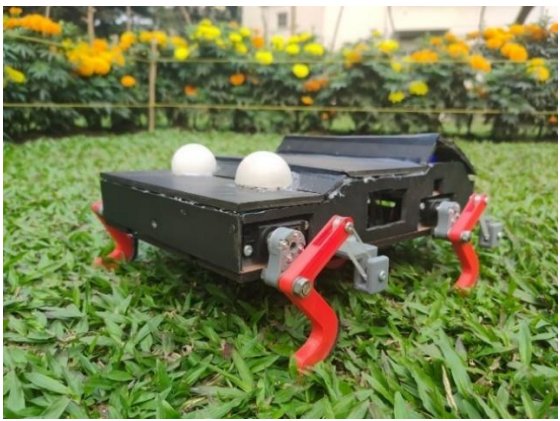

**Fig.1** GiAnt the bio-inspired hexapod.

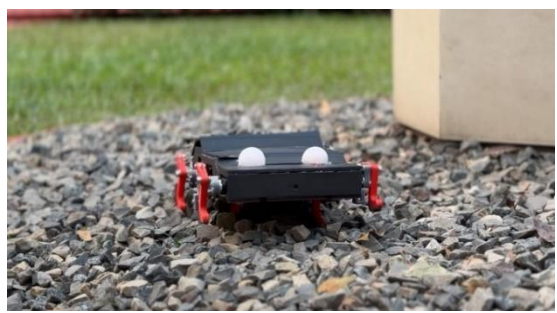

**Fig.2** GiAnt traversing through rocky terrain.

**2.Methodology**
2.1 Mechanical design

The 3D modeling of GiAnt was completed using SOLIDWORKS 2022. GiAnt is designed with forward, backward, left and right motion capabilities. The dimensions are as follows: length 310 mm, width 200 mm, height 120 mm.

2.1.1 Leg

GiAnt has a unique leg design (Fig.3) incorporating a link and a crank. The link enables better interaction with the ground, and the crank gives the luxury of ascending upper platforms from the ground. Each of the legs are operated using a 10 kgf/cm servo motor. Fitting it's six legs in a short dimension was challenging, but finally the base was kept to 310 mm.

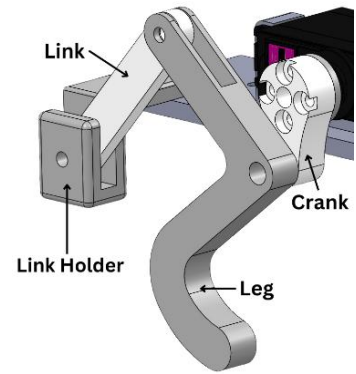

**Fig.3** 3D model of leg.

2.1.2 Assembly version 1

In the first version of assembly (Fig.4), the middle legs were placed inside to avoid collisions with the front and rear legs, keeping the servo motors outward. This version of assembly was unsuccessful because of the thin portions beside the servo motor holder, where the stress was concentrated (Fig.5).

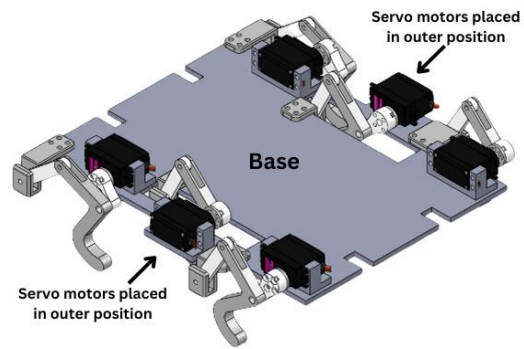

**Fig.4** 3D model of assembly version 1.

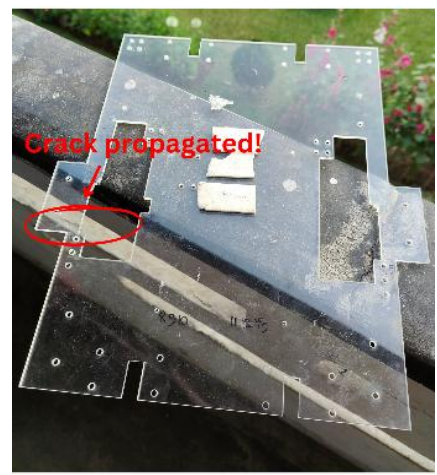

**Fig.5** Crack propagation in base of assembly version 1.

### 2.1.3 Assembly version 2

In this version of assembly, the middle servo motors are shifted inward, keeping the legs in the same position (Fig.6). Thus, there is no thin portion in the base, and it is stronger than the previous version. Therefore, manufacturing was started.

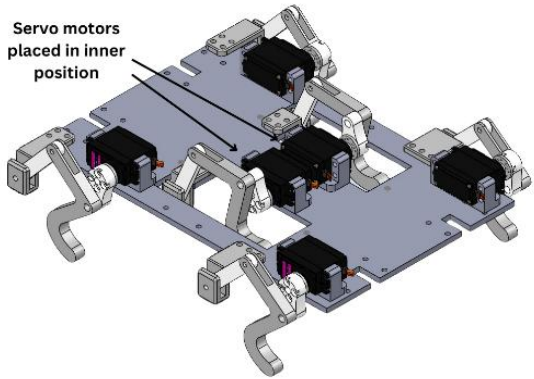

**Fig.6** 3D model of assembly version 2

### 2.2 Structural Analysis

The finalized CAD model was analyzed by finite element analysis using ANSYS Mechanical. While analyzing the leg, the material properties were considered according to PLA. 1.5 kg-wt load was applied on each leg, considering factor of safety 3. The results from FEA (finite element analysis) are shown below:

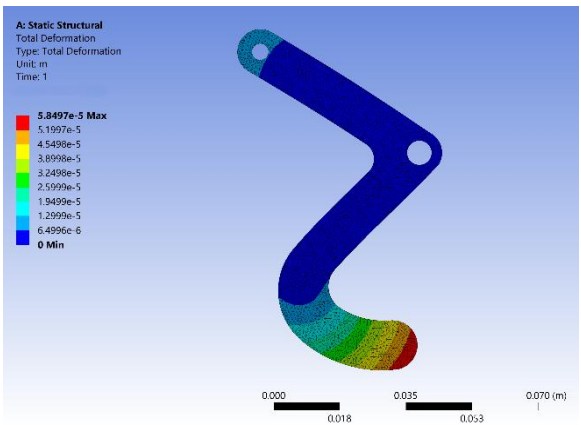

**Fig.7** Deformation in the leg.

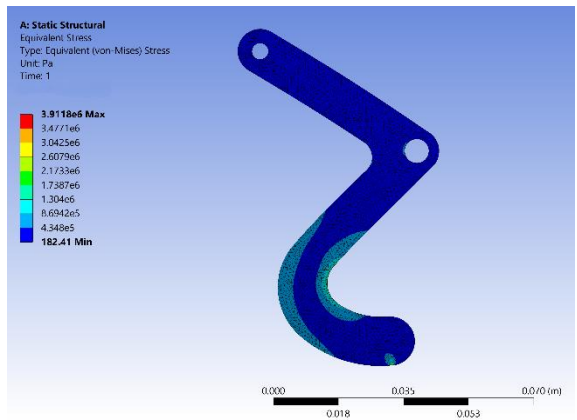

**Fig.8** Developed stress in the leg.

The stress and deformation analysis results were found to be satisfactory, and hence the manufacturing process was started.

### 2.3 Manufacturing process

The base of the GiAnt was first manufactured from acrylic sheets, but that didn't last because of its brittle nature. Then the base was manufactured using 6 mm plywood sheet, which was quite rigid and sustainable. Links used in the legs are also made from acrylic sheets, and other parts of the leg are 3D printed using PLA (polylactic acid). The outer shell is made from PVC sheets.

### 2.3.2 Laser cutting

The base and the links are laser cut from plywood and acrylic sheets, respectively (Fig.7).

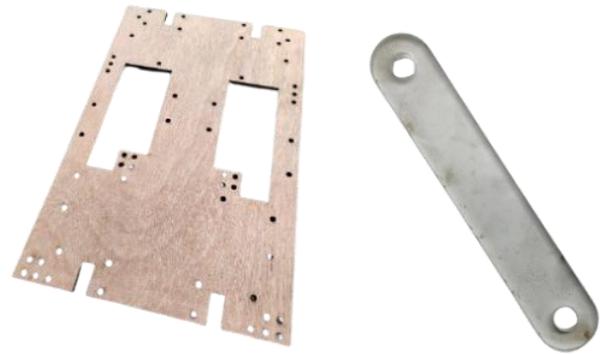

**Fig.9** Laser cut parts.

### 2.3.3 3D printing

The parts of the leg excluding the link were manufactured by 3D printing (Fig.8).

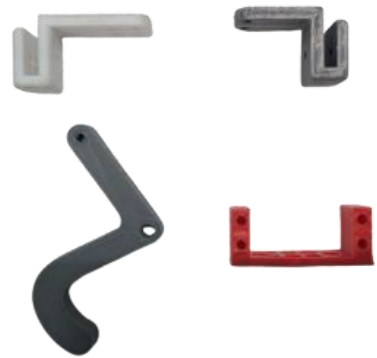

**Fig.10** 3D printed parts.

### 2.3.4 Handicraft

The outer shell is constructed by shaping a PVC sheet through manual cutting.

### 2.4 Gait analysis

To determine a leg's ground-touching point trajectory, SOLIDWORKS Motion Analysis is used. After getting the coordinates from the SOLIDWORKS Animation, the curve was fitted using the plotting software OriginPro. The curve (Fig.9), representing the leg's ground-touching point trajectory, was found through the Savitzy-Golay method of Signal Processing. The smooth trajectory confirms the improved interaction with the ground.

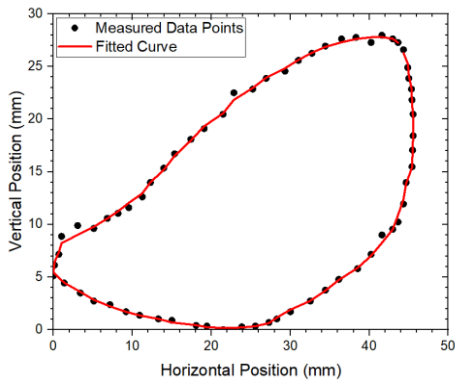

**Fig.11** Leg's ground-touching point trajectory.

2.5 Electrical system

We have used Arduino as our main electronic component for running the system. Printed circuit board is used to connect all components effortlessly and easily (Fig.10). Other main components are listed below:

**Table.1** Electrical components and purposes

| Electrical Components | Purpose |
| --- | --- |
| Arduino Mega | Provides multiple PWM pins to control servo motors efficiently, eliminating the need for additional drivers. |
| Servo Motor (MG996R) | Provides full 360° rotation needed to move the robot's legs |
| Buck Converter (LM2596) | Provides safe voltage levels for components by lowering the 12V battery output to 7.2V and 5V as needed. |
| Bluetooth Module (HC-06) | Enables wireless control, allowing the robot to be operated remotely. |
| ESP32-CAM | Captures live video and performs real-time object recognition to enhance robot navigation. |

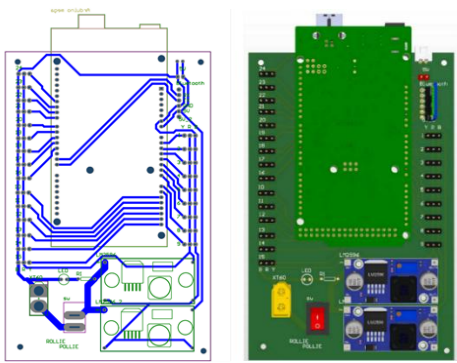

**Fig.12** Printed circuit board schematic.

PCB design was made with Altium software.

2.6 Software system

The software system of the GiAnt is designed to facilitate real-time control, image processing, and object detection, making it adaptable to various tasks. Key software tools and frameworks used include Arduino IDE, Python and OpenCV for image processing, alongside the YOLO object detection algorithm.

2.6.1 Control system and communication

The robot is controlled through a custom Bluetooth control application named "Bluetooth Electronics" (Version 1.5) (Fig.13), allowing remote operation. The buttons of UI were manually customized to perform specific functions. Such as the leftmost 8 directional buttons are used to provide forward, backward and diagonal motion, the numerical 1, 2, 3 buttons are used to move the 1st, 2nd and 3rd pairs of legs together respectively, the rightmost 6 pairs of arrow buttons are used to control individual leg movement and the red and green buttons are used to control the priming of the two sets of triangularly synchronized legs. Using an Arduino Mega microcontroller, the control system manages the coordinated movement of the robot's legs in pairs, allowing for stable and agile navigation on uneven terrains. Each button sends a unique alphabetic signal to the HC-06 Bluetooth module which is interpreted by the Arduino to execute the corresponding command. Each leg movement is precisely synchronized through pulse-width modulation (PWM) signals generated by the Arduino.

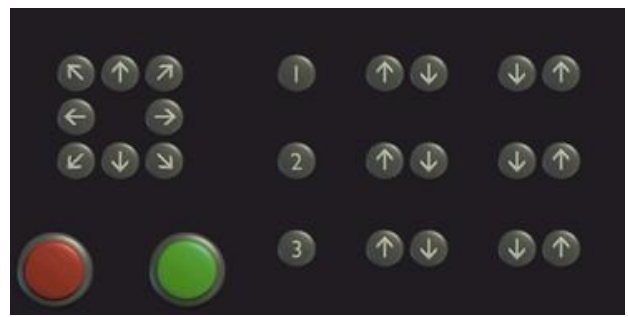

**Fig.13** Customized bluetooth control application.

2.6.2 Image processing and object Detection

Image processing and object detection are core functions implemented on the ESP32-CAM module, which provides a live video stream that is essential for visual monitoring and object recognition. The ESP32-CAM captures video frames and transmits them to a PC over a Wi-Fi connection, where they are processed in real time using OpenCV, a widely used computer vision library. For object detection, YOLOv3, known for its speed and accuracy, identifies objects by applying bounding boxes within the video feed. This setup enables the robot to recognize and label obstacles, which is particularly valuable in applications like search and rescue where quick identification and maneuverability are critical. To address the resolution limitation of the ESP32-CAM, drone cams and Z-cams can be considered as alternatives. Drone cameras provide higher resolution and advanced stabilization features, which can significantly enhance the detection accuracy. Moreover, Z-cams utilize depth-sensing technology, which can provide 3D data to improve object detection, enabling more precise recognition during movement.

The combined use of real-time control, image processing, and object detection allows the GiAnt robot to interact with its environment intelligently, making it suitable for tasks such as surveillance, exploration, and assistance in hazardous conditions.

## 3. Results and discussions

After completion of manufacturing, test runs were made. After a few trials adjustments were made and run again. Repeating this process, better results were achieved.

3.1 Motion capabilities

All the motions are ensured using a 360-degree continuous rotation servo. The direction of rotation and speed of rotation are controlled to implement various types of motion.

3.1.1 Forward and backward motion

Two sets of legs were introduced as triangularly synchronized. When one set of legs is touching the ground, the other set of legs is making a phase difference of 90 degrees with the previous set. That means any of the set of three legs is in contact with the ground simultaneously. The backward motion was obtained by altering the direction of the rotation of each leg.

3.1.2 Left and right rotation

When GiAnt was turning left and right, the three legs on one side rotated in such a way that others rotated in the opposite direction.

3.1.3 Elevation from the ground

The curve (Fig.14) illustrates the position of legs above the ground with respect to time. The data points of the curve were extracted from the SOLIDWORKS Motion Analysis and the curve was fitted using Polynomial Fit in plotting software OriginPro. The equation of the curve from Polynomial Fit was shown as below:

$y = 27.66182 - 1.67994 x^1 - 0.00283 x^2 + 0.00147 x^3 - (1.59275 \times 10^{-5}) x^4$

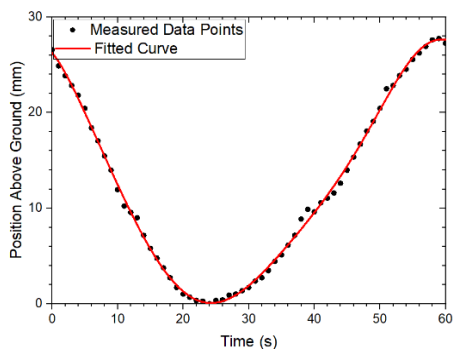

**Fig.14** Position of leg above ground with respect to time.

Here from the curve (Fig.14), it is observed that a single leg cannot be elevated more than 2.7 cm above the ground, but when the legs are operated in such a systemic way that the rear legs are in the lowest position and the middle legs are in the highest position, the maximum possible elevation can be up to 8 cm (Fig.15).

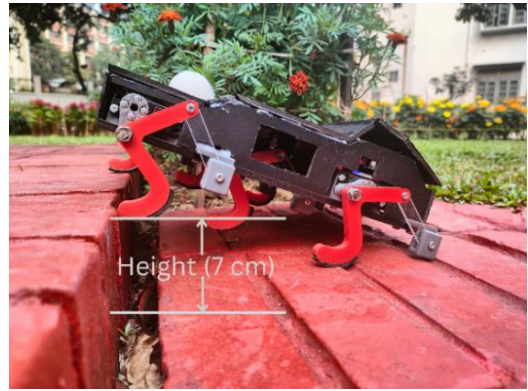

**Fig.15** GiAnt ascending a 7 cm pavement

3.2 Object detection capabilities

The ESP32-CAM module provides real-time video streaming, enabling GiAnt to detect and label obstacles using OpenCV and YOLOv3 (Fig.14, Fig.15). This lets the robot react quickly to its surroundings, making it better for tasks like search and rescue. The system ensures GiAnt can move and adapt intelligently in challenging environments.

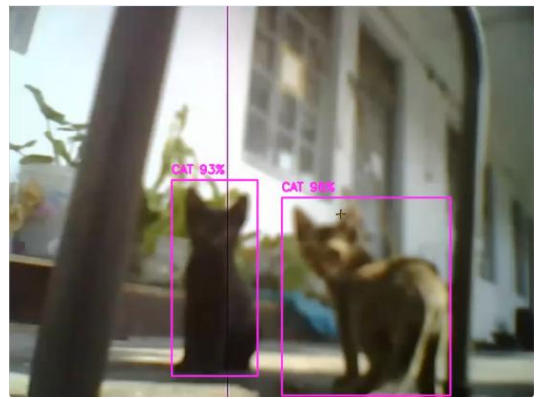

**Fig.16** GiAnt detecting two cats in a single frame

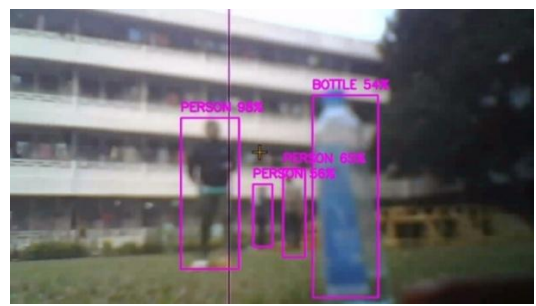

**Fig.17** GiAnt detecting various kinds of objects.

3.3 Performance capabilities of GiAnt

The table below summarizes GiAnt's performance across key metrics, including its climbing capabilities, terrain navigation, and load performance. These metrics highlight the robot's adaptability and potential for improvement with configuration changes.

Table.2 Summary of Giant's performance metrics

| Feature | Performance Details |
|---|---|
| **Climbing Capability** | a. Easily climbs heights of 2 to 5 cm<br>b. Handles 5 to 8 cm heights with some difficulty<br>c. Heights greater than 8 cm are not possible for this size |
| **Terrain Navigation** | a. Walks easily on grass and stones/pebbles of 2.5 to 3 cm diameter<br>b. Experiences difficulty with pebbles greater than 3 cm in diameter |
| **Load Performance** | a. Current configuration supports loads of 1.2–2.5 kg<br>b. Replacing servo motors with 20 kg/cm variants increases load capacity to 5 kg |

## 4. Limitations and future improvements

- ESP32 camera's low resolution. FPV cameras could be employed.
- Use of high-torque motors is recommended for improved efficiency.
- More performance could be achieved with a 360-degree continuous servo that provides position feedback
- Optimized gait control algorithms, and autonomy, could further expand GiAnt's capabilities.

## 5. Conclusion

In conclusion, this work outlines GiAnt, an affordable biomechanically inspired hexapod robot that has been successfully designed and developed with object detection capability. The following are the key findings from this work:

1) The robot features a lightweight, affordable structure weighing 1.75 kg with dimensions of 310 mm x 200 mm x 120 mm.
2) GiAnt employs a link and crank mechanism in its leg design to ensure effective interaction with ground and stable motion.
3) The robot has a unique capability to ascend up to 8 cm from the ground, which is a key achievement of this project.
4) GiAnt successfully walks through unsmooth paths like rocks and grass, which makes it suitable exploring bot for a wide range of terrain.
5) The robot demonstrates a capability of real-time object detection, which would provide support during operation in challenging terrain and remote monitoring.

Finally, it is evident that the GiAnt is expert in traversing through irregular and steep terrain with ease. Since it is equipped with a real-time live monitoring system, it can be used for search and rescue operations, exploring hazardous environments that are not suitable for direct human contact, package delivery, agricultural monitoring, assisting people with disabilities, etc. Though the hexapod isn't developed to the fullest, a few modifications can bring about massive upgrades, which can be a great tool in the modern advancement of hexapodal robotics.

## 6. Acknowledgement

The authors would like to express their sincere gratitude to the esteemed course instructors: Sakib Javed, Lecturer; Mr. Shahriar Alam, Lecturer; Mr. Priom Das, Lecturer; and Dr. Kazi Arafat Rahman, Associate Professor, all from the Department of Mechanical Engineering at the Bangladesh University of Engineering and Technology. Their invaluable guidance was instrumental in the successful completion of this research, greatly enhancing the quality of this conference paper. Their insightful support throughout the process has helped a lot.

**NOMENCLATURE**

*DOF: Degress of freedom*
*PCB: Printed circuit board*
*3D: 3 Dimensional*
*FEA: Finite element analysis*
*PLA: Polylactic acid*